\documentclass[11pt]{scrartcl}
\usepackage{fullpage}
\usepackage{hyperref}
\usepackage{microtype}
\usepackage{xurl}
\usepackage{tikz}
\usetikzlibrary{positioning, shapes.geometric, arrows.meta, calc, intersections}
\usepackage{graphicx}
\usepackage{wrapfig}
\usepackage{subcaption}
\usepackage[dvipsnames]{xcolor}

\usepackage{color-edits} %[suppress] [showdeletions]
\addauthor{mc}{Mulberry} % creates automatic commands \mcedit{}, \mccomment{}, \mcdelete{}
\addauthor{ig}{TealBlue}

\definecolor{brandYellow}{HTML}{FFC800}

\newcommand{\ratingimg}[1]{\includegraphics[width=1.6cm]{#1}}

\tikzset{
  decision/.style={
    diamond,
    draw,
    fill=brandYellow,
    align=center,
    text width=4cm,
    minimum width=5cm,
    shape aspect=1.618,
    inner sep=-8pt
  },
  line/.style={
    line cap=round,
    line width=1pt
  },
  arrow/.style={
    line,
    -Stealth
  },
  label/.style={
    font=\footnotesize\bfseries,
    inner sep=2pt
  },
  rating/.style={
    inner sep=0pt
  }
}

\title{The Reported Engagement with AI Level (REAL) Rating: A Framework for Disclosing Human-AI Collaboration}
\author{\scalebox{0.97}[1.0]{Im\`ene Goumiri, Mayleen Cortez-Rodriguez, Eric Bell, Amanda Muyskens}}
\date{Real Good AI}

\begin{document}

\maketitle

\begin{abstract}
Generative artificial intelligence has become increasingly incorporated into digital media and more generally into production workflows with which the public frequently interacts. Current provenance standards and disclosure methods frequently rely on binary categorizations, differentiating only between entirely human-authored and AI-generated content, or depend on technical watermarking that lacks user-facing clarity. However, there are very different risks and outcomes depending on the different uses of AI in the creation and consumption of end products. Therefore, we propose the Reported Engagement with AI Level (REAL) Rating framework which addresses these concerns by introducing a six-tier scale (Levels 0-5) that discloses the extent of AI involvement in both product creation and user experience. This paper details the structural methodology of the REAL Rating system and the specifics of its application across text, image, audio, video, and software products.
\end{abstract}

\section{Introduction}

Generative artificial intelligence (AI) has rapidly proliferated across the digital landscape, fundamentally altering media and product development. Recent industry analyses indicate that up to 74\% of newly published web pages contain some form of AI-generated content~\cite{whitehat2026marketing}. Without established provenance standards, this surge in synthetic media creates significant challenges for information integrity, as high-fidelity AI outputs, such as deepfakes and automated text, become increasingly difficult for audiences to distinguish from human-authored content~\cite{pai2026responsible}. The implications of this authenticity crisis extend to academic integrity, institutional trust, and legal compliance, necessitating reliable methods for disclosing AI involvement~\cite{gunderson2026ailaws}.

A primary driver for formal disclosure is the technical nature of Large Language Models (LLMs) and other generative systems, which are prone to the generation of factually incorrect or nonsensical information presented with high confidence, referred to as ``hallucinations''~\cite{ji2023hallucination}. Because these systems function as probabilistic predictors rather than knowledge bases, the inclusion of AI in a workflow introduces specific reliability risks that differ from human error. Precise disclosure allows the end-user to apply appropriate levels of scrutiny based on the known involvement of these systems.

Public sentiment often fluctuates between total rejection of synthetic media and a misunderstanding of non-generative AI tools. Many useful applications of machine learning, such as automated noise reduction or basic spelling correction, are frequently conflated with fully generative LLM outputs because they are also called AI tools. This lack of distinction can lead to a complete dismissal of products that utilize any AI, even when its use does not actually compromise human authorship. By providing a granular scale, creators can acknowledge the use of helpful tools without being mischaracterized as having abdicated the creative process to an algorithm.

Several distinct approaches have emerged to address AI disclosure, each serving different operational interests and presenting unique limitations. The first category comprises automated AI detection tools, which analyze media to predict the probability of algorithmic generation. While widely adopted in educational and publishing sectors, these tools suffer from significant reliability issues, including high false-positive rates that disproportionately affect non-native English speakers~\cite{weberwulff2023detection}. Furthermore, studies demonstrate that sophisticated AI-generated content frequently evades detection entirely, rendering these systems inadequate as standalone verification methods~\cite{weberwulff2023detection}.

A second category centers on cryptographic watermarking and provenance metadata, such as the widely adopted Coalition for Content Provenance and Authenticity (C2PA) standard. These systems embed hidden digital signatures into files at the point of creation, providing a verifiable chain of custody~\cite{microsoft2026authenticity}. This approach primarily serves the liability and compliance interests of major technology companies and hardware manufacturers. However, metadata is fragile; it can be stripped during file compression, format conversion, or intentional tampering, and invisible watermarks fail to provide immediate, transparent disclosure to the end-user without specialized software~\cite{microsoft2026authenticity}.
Further, this is extremely domain-dependent and is not general enough to deal with multi-media products such as plain-text documents, transcoded social media clips, generative multi-modal composites, or static screenshots.

A third approach involves platform-level, binary labeling (e.g., ``Made with AI" tags applied by social media networks). While offering user-facing visibility, these labels primarily serve platform compliance and public relations more than precise provenance, as they rely on a rigid categorization that fails to capture the realities and nuances of modern digital workflows~\cite{pai2026responsible, crootof2023humans}. Contemporary content creation frequently involves a ``human-in-the-loop" paradigm, utilizing AI for intermediate tasks such as brainstorming, grammar correction, or peripheral image adjustments~\cite{crootof2023humans}.

Addressing these disclosure issues is not possible with current technology alone. 
Existing disclosure methods like detection algorithms, fragile metadata, or binary platform tags are ill-equipped to communicate this spectrum of collaboration.
Therefore, we propose the \textbf{Reported Engagement with AI Level (REAL) Rating}, a framework which addresses this critical gap by providing a voluntary nuanced, standardized vocabulary for declaring the specific degree of human-AI interaction, replacing inadequate binary assessments with a graduated scale of engagement. In doing so, we appeal both to the integrity of humanity and to public pressure to produce a general, widely applicable solution. Although this may seem optimistic, it is commonplace already in terms of plagiarism, authorship, and other similar issues. 

The REAL Rating approach draws a conceptual parallel to the SAE J3016 levels of driving automation, which range from no automation (Level 0) to full automation (Level 5)~\cite{on2021taxonomy}. Just as the automotive industry moved away from the binary concept of ``self-driving'' to specify the level of human intervention required, the REAL Rating framework acknowledges that digital content creation exists on a similar spectrum. Defining these levels allows for a shared understanding of responsibility and agency between the human creator and the generative system.

In this manuscript, we establish the REAL 0-5 scale definitions, give rating and marking methodology, and also provide a few examples of implementation.

\section{The REAL Rating Framework}

The REAL Rating system quantifies AI involvement on a scale from 0 to 5. Each level represents a distinct threshold of machine contribution relative to human authorship. These levels are meant to be general for simple adoption across media such as written communication, images, videos, audio, more general products, etc. Therefore, the descriptions are very general, but conceptually represent the amount of AI incorporation in the creation. 
Moreover, these levels are not meant to be a judgment on the creators, but rather a disclosure to the public of the AI involved in the creation of content or the AI consumers may interact with while using a product. We define the levels below with some simple examples. Later, we provide more concrete, detailed examples to illustrate the levels.

\begin{description}

\item[Level 0 (No AI):] The content was created entirely by a human without the use of AI tools.

\item[Level 1 (Automation):] AI was utilized strictly for automated refinement or basic tooling, but the core ideas, primary structure, and presentation remain entirely human-generated. Examples include using tools like spell-check, grammar correction (e.g. Grammarly), automated audio cleanup, or automated green-screen/cutout features in video editing. 

\item[Level 2 (Brainstorming):] AI was used for ideation, structural organization, or consultation, but the final output is written or constructed by the human author. Examples include using Large Language Models (LLMs) to generate outlines, brainstorm concepts, or summarize external information.

\item[Level 3 (Moderate AI):] Generative AI components are present within the work, but they do not constitute the primary focus nor do they alter the core meaning of the work. This includes incorporating underlying components previously rated at Level 3 or, for example, the use of generative fill on peripheral image elements.

\item[Level 4 (Major AI):] AI contributions are substantial. In text, this applies if the AI were considered equivalent to a human co-author. In visual media, this applies if the generated or traced portion is unique, central, or constitutes a large part of the final image or video. This also applies if AI substantially changes the meaning, context, or primary object of the media.

\item[Level 5 (Full AI):] The media was generated entirely from prompts (e.g., using Sora for video or Midjourney for images), or an entire document was copied from an LLM, regardless of subsequent human editing. In product and research environments, this level also applies to platforms where end-users interact directly with AI models.

\end{description}

A critical function of these levels is to demonstrate increasing risk. As the rating moves from Level 1 to Level 5, the opportunity of algorithmic bias, hallucination, and loss of human agency increases, requiring higher levels of consumer awareness.

\subsection{Markings and implementation}

\begin{figure}
    \centering
    \ratingimg{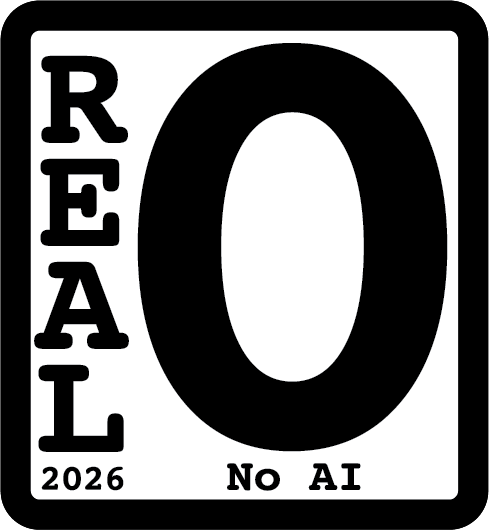}\hspace{5mm}%
    \ratingimg{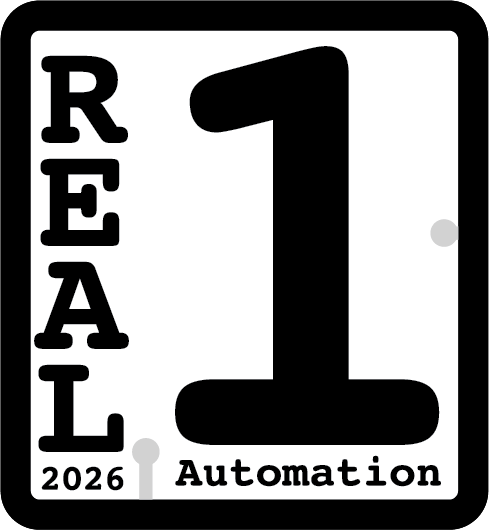}\hspace{5mm}%
    \ratingimg{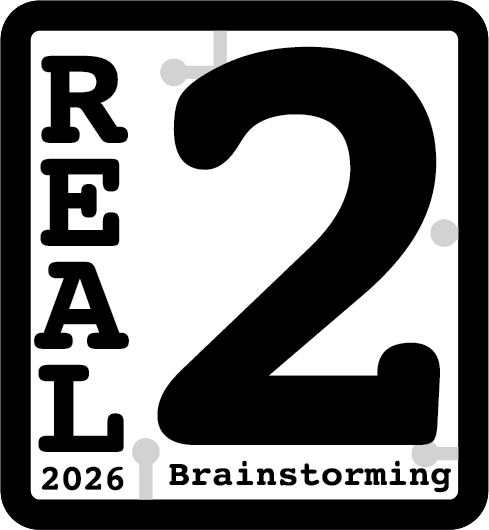}\hspace{5mm}%
    \ratingimg{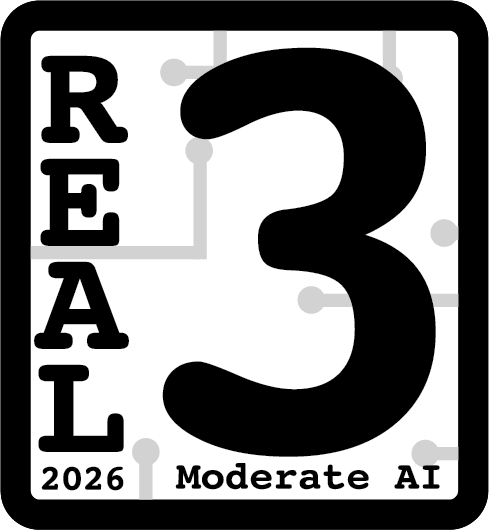}\hspace{5mm}%
    \ratingimg{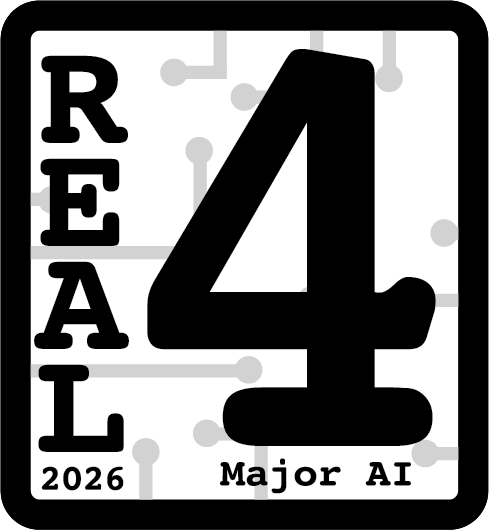}\hspace{5mm}%
    \ratingimg{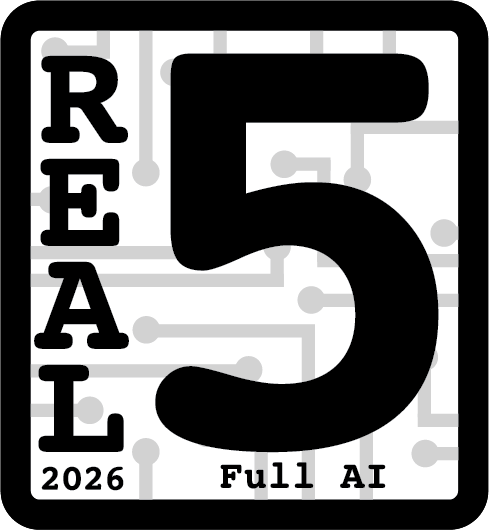}
    \caption{REAL rating markings.}
    \label{fig:markings}
\end{figure}

To facilitate immediate and intuitive recognition of AI involvement, we have developed a standardized set of visual identifiers for each level of the framework (\autoref{fig:markings}). These markings are released into the public domain to encourage broad, unrestricted adoption across all creative and technical sectors.

The visual system is designed for high visibility across diverse digital and physical environments. For maximum efficacy, these identifiers should be placed prominently within the user's primary field of view:
\begin{description}
    \item[Documents:] Located on the title page or header.
    \item[Visual Media:] Positioned in the least intrusive corner of the image or as an overlay on video/song thumbnails. Marking needs to be present only for 3 seconds at the beginning of a posted the video, but should remain on screen if the video is live and viewers may join at any point in the broadcast. 
    \item[Communications:] Integrated into footers, such as in professional email correspondence or website design. 
\end{description}

\begin{figure}
    \centering
    \begin{subfigure}{0.9\linewidth}
        \centering
        \includegraphics[width=\linewidth]{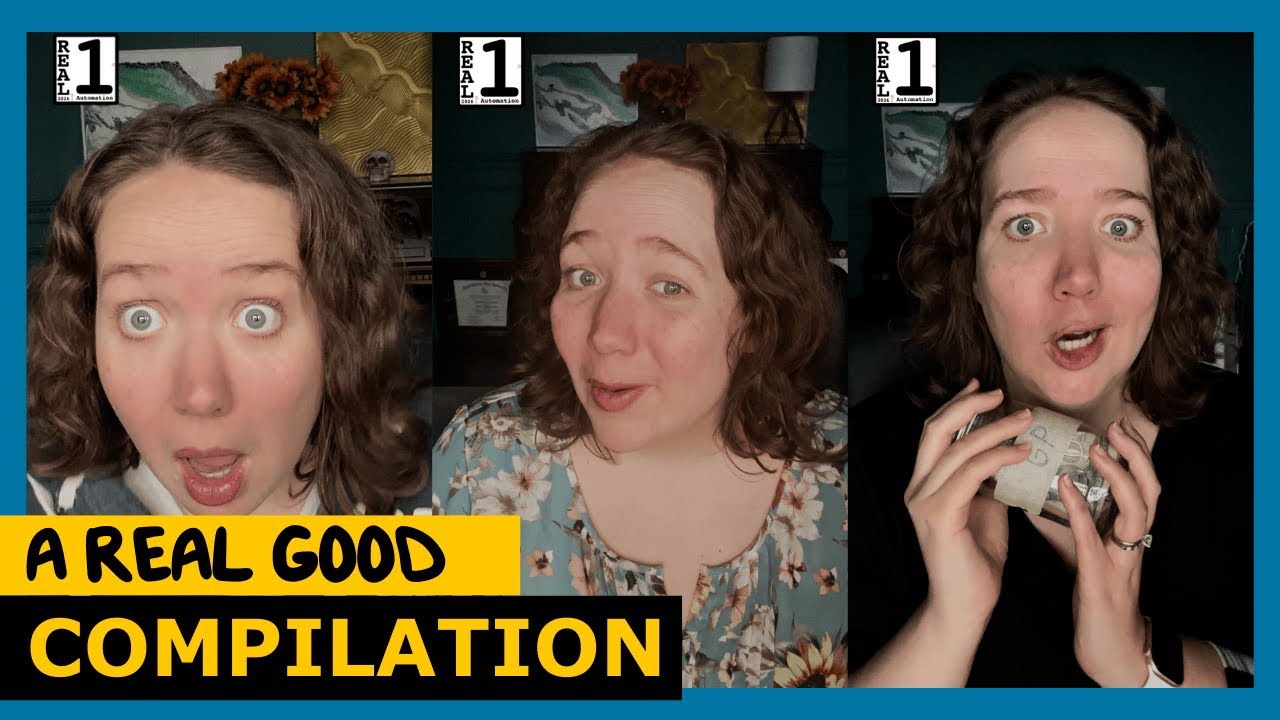}
        \caption{Video thumbnails.}
    \end{subfigure}\\[1em]
    \begin{subfigure}{0.45\linewidth}
        \centering
        \includegraphics[width=\linewidth]{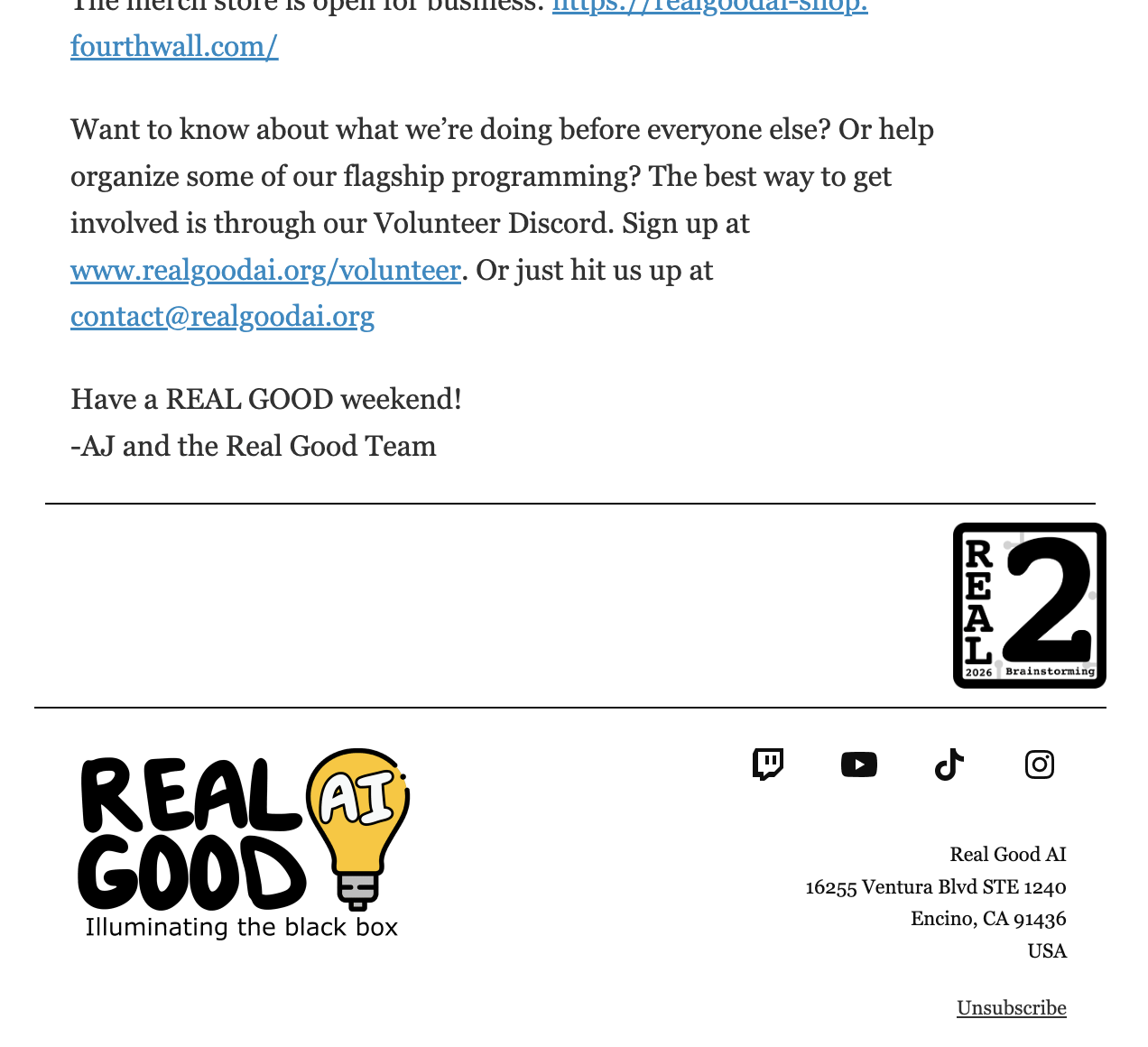}
        \caption{Email footer.}
    \end{subfigure}\hspace{0.05\linewidth}%
    \begin{subfigure}{0.4\linewidth}
        \centering
        \includegraphics[width=\linewidth]{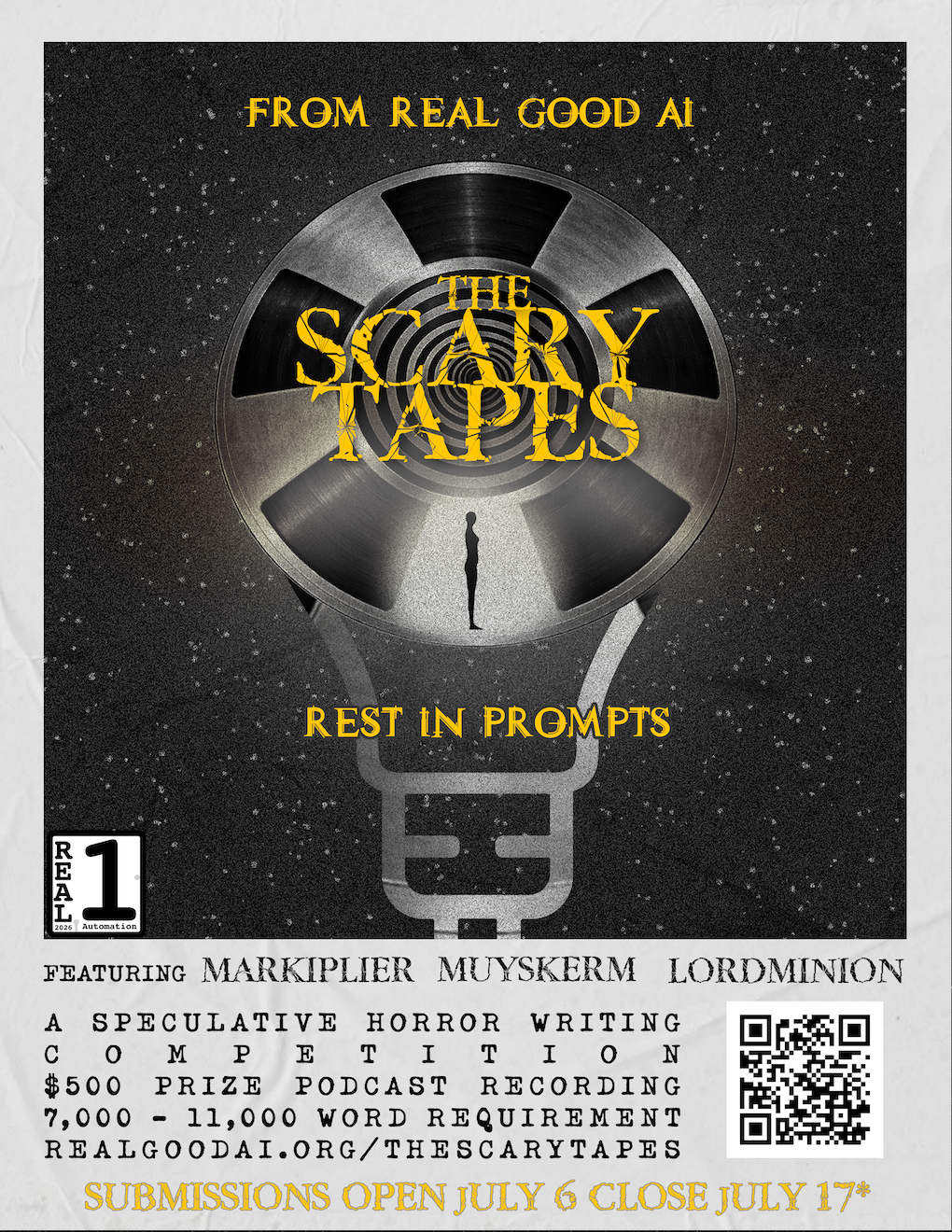}
        \caption{Poster.}
    \end{subfigure}
    \caption{Examples of REAL rating markings applied to various media.}
    \label{fig:examples}
\end{figure}

Examples of these applications across various media formats are illustrated in \autoref{fig:examples}.

Further, the aspect ratios, colorings, and other identifying features of the markings should remain unchanged.

\section{System Application and Methodology}

The REAL Rating framework utilizes a branching decision-tree methodology to determine the appropriate disclosure level. Creators evaluate their work through a progressive series of binary logic gates. Because modern media frequently combine multiple types of content, the framework isolates evaluation into three primary modalities: visual, audio, and text.

To determine the overall rating for a multimodal product, such as a video, podcast, or illustrated blog post, the creator evaluates each constituent modality independently using its corresponding flowchart. The final REAL Rating applied to the entire work is the highest level achieved across any evaluated modality. For example, a video game featuring entirely human-written code (Level 0) and human-created 3D models (Level 0) but utilizing AI-generated background music (Level 4) receives an overall rating of Level 4.

The evaluation process for each modality begins with primary triage questions that assess the foundational role of AI in the project, such as whether a generative model was consulted for ideation or if synthetic elements were present in the source materials. These initial nodes route the evaluation into either a low-involvement track (Levels 0 and 1) or a high-involvement track (Levels 2 through 5). Subsequent criteria questions further isolate specific workflows to assign the final rating for that modality.

\subsection{Visual Evaluation}

\begin{figure}[h!tbp]
  \centering
  \begin{tikzpicture}[x=5cm, y=4cm]
    \node[decision] (A) at (0, -0.3) {Did you \\ use generative AI for inspiration or \\ prompting?};
    \node[decision] (B) at (-1,-1) {Was any \\ of your source material AI-\\generated?};
    \node[decision, name path=C] (C) at (-1,-2.5) {Did you \\ use automated tools for cleanup or \\ editing?};
    \node[decision] (D) at (1,-1) {Did you \\ generate or ``fill'' any part of the \\ visuals?};
    \node[decision] (E) at (1,-2) {Was the \\ generated portion a central part of the \\ visuals?};
    \node[decision, name path=F] (F) at (1,-3) {Are any \\ portions of the visuals entirely human-\\made?};
    
    \node[rating] (R0) at (-1.25,-3.8) {\ratingimg{images/0.png}};
    \node[rating] (R1) at (-0.75,-3.8) {\ratingimg{images/1.png}};
    \node[rating] (R2) at (-0.25,-3.8) {\ratingimg{images/2.png}};
    \node[rating] (R3) at (0.25,-3.8) {\ratingimg{images/3.png}};
    \node[rating] (R4) at (0.75,-3.8) {\ratingimg{images/4.png}};
    \node[rating] (R5) at (1.25,-3.8) {\ratingimg{images/5.png}};
    
    \draw[arrow] (A) -| node[label, pos=0, anchor=south east] {No} (B);
    \draw[arrow] (A) -| node[label, pos=0, anchor=south west] {Yes} (D);
    \draw[arrow] (B) -- node[label, pos=0, anchor=south west] {Yes} (D);
    \draw[arrow] (B) -- node[label, pos=0, anchor=north west] {No} (C);
    \path[name path=V0] (C -| R0) -- (R0);
    \draw[arrow, name intersections={of=C and V0}] (intersection-1) -| node[label, pos=0, anchor=north east] {No} (R0);
    \path[name path=V1] (C -| R1) -- (R1);
    \draw[arrow, name intersections={of=C and V1}] (intersection-1) -| node[label, pos=0, anchor=north west] {Yes} (R1);
    \draw[arrow] (D.240) -| node[label, pos=0.05, anchor=south east] {No} (R2);
    \draw[arrow] (D) -- node[label, pos=0, anchor=north west] {Yes} (E);
    \draw[arrow] (E) -- node[label, pos=0, anchor=north west] {Yes} (F);
    \draw[arrow] (E) -| node[label, pos=0, anchor=south east] {No} (R3);
    \path[name path=V4] (F -| R4) -- (R4);
    \draw[arrow, name intersections={of=F and V4}] (intersection-1) -| node[label, pos=0, anchor=north east] {Yes} (R4);
    \path[name path=V5] (F -| R5) -- (R5);
    \draw[arrow, name intersections={of=F and V5}] (intersection-1) -| node[label, pos=0, anchor=north west] {No} (R5);
  \end{tikzpicture}
  \caption{Visual evaluation flowchart}\label{fig:visual}
\end{figure}
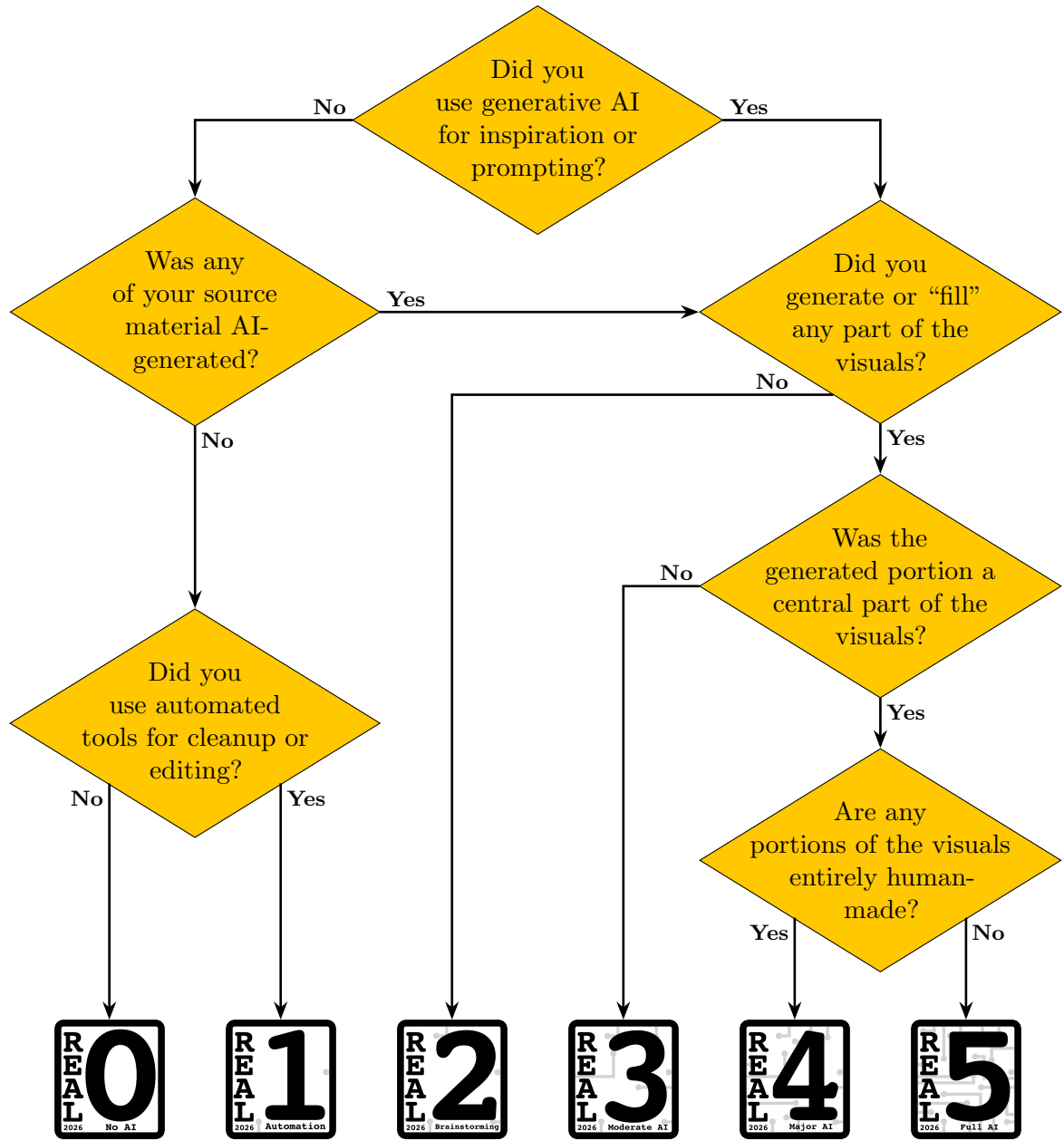

Visual evaluation differentiates between fully generated imagery, central generative modifications, peripheral generative fill, AI-assisted ideation, and basic automated adjustments. This modality covers static images, physical product designs, user interfaces, and the visual components of video production (\autoref{fig:visual}).

\subsection{Audio Evaluation}

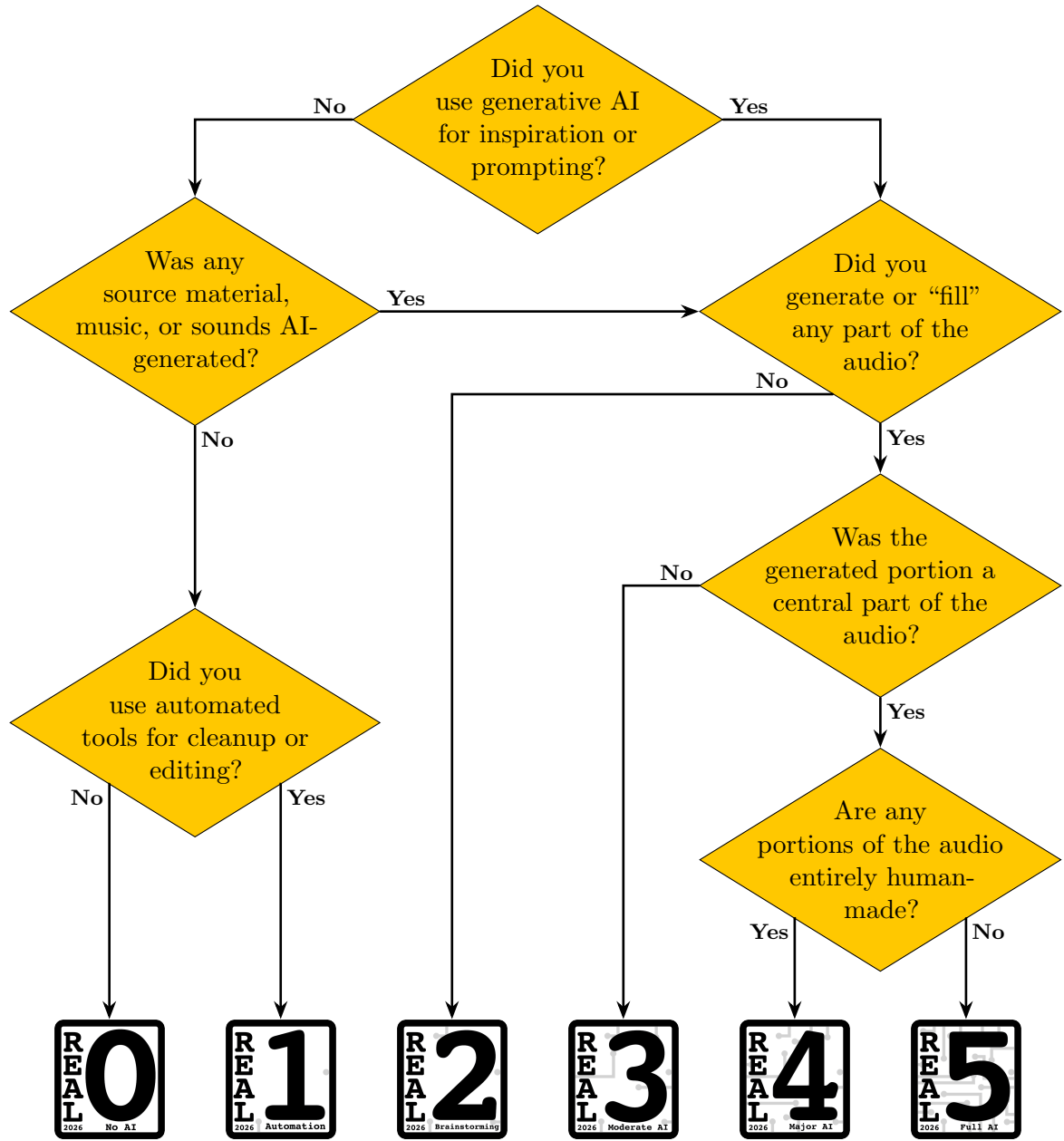
\begin{figure}[h!tbp]
  \centering
  \begin{tikzpicture}[x=5cm, y=4cm]
    \node[decision] (A) at (0, -0.3) {Did you \\ use generative AI for inspiration or \\ prompting?};
    \node[decision] (B) at (-1,-1) {Was any \\ source material, \\ music, or sounds AI-\\generated?};
    \node[decision, name path=C] (C) at (-1,-2.5) {Did you \\ use automated tools for cleanup or \\ editing?};
    \node[decision] (D) at (1,-1) {Did you \\ generate or ``fill'' any part of the \\ audio?};
    \node[decision] (E) at (1,-2) {Was the \\ generated portion a central part of the \\ audio?};
    \node[decision, name path=F] (F) at (1,-3) {Are any \\ portions of the audio entirely human-\\made?};
    
    \node[rating] (R0) at (-1.25,-3.8) {\ratingimg{images/0.png}};
    \node[rating] (R1) at (-0.75,-3.8) {\ratingimg{images/1.png}};
    \node[rating] (R2) at (-0.25,-3.8) {\ratingimg{images/2.png}};
    \node[rating] (R3) at (0.25,-3.8) {\ratingimg{images/3.png}};
    \node[rating] (R4) at (0.75,-3.8) {\ratingimg{images/4.png}};
    \node[rating] (R5) at (1.25,-3.8) {\ratingimg{images/5.png}};
    
    \draw[arrow] (A) -| node[label, pos=0, anchor=south east] {No} (B);
    \draw[arrow] (A) -| node[label, pos=0, anchor=south west] {Yes} (D);
    \draw[arrow] (B) -- node[label, pos=0, anchor=south west] {Yes} (D);
    \draw[arrow] (B) -- node[label, pos=0, anchor=north west] {No} (C);
    \path[name path=V0] (C -| R0) -- (R0);
    \draw[arrow, name intersections={of=C and V0}] (intersection-1) -| node[label, pos=0, anchor=north east] {No} (R0);
    \path[name path=V1] (C -| R1) -- (R1);
    \draw[arrow, name intersections={of=C and V1}] (intersection-1) -| node[label, pos=0, anchor=north west] {Yes} (R1);
    \draw[arrow] (D.240) -| node[label, pos=0.05, anchor=south east] {No} (R2);
    \draw[arrow] (D) -- node[label, pos=0, anchor=north west] {Yes} (E);
    \draw[arrow] (E) -- node[label, pos=0, anchor=north west] {Yes} (F);
    \draw[arrow] (E) -| node[label, pos=0, anchor=south east] {No} (R3);
    \path[name path=V4] (F -| R4) -- (R4);
    \draw[arrow, name intersections={of=F and V4}] (intersection-1) -| node[label, pos=0, anchor=north east] {Yes} (R4);
    \path[name path=V5] (F -| R5) -- (R5);
    \draw[arrow, name intersections={of=F and V5}] (intersection-1) -| node[label, pos=0, anchor=north west] {No} (R5);
  \end{tikzpicture}
  \caption{Audio evaluation flowchart}\label{fig:audio}
\end{figure}

Audio evaluation categorizes the use of AI voice generation, synthetic music, generated sound effects, and automated audio cleaning tools. This assessment applies to independent audio tracks as well as the sound components of broader multimedia projects, separating the generation of the audio itself from the drafting of the underlying script or lyrics (\autoref{fig:audio}).

\subsection{Text Evaluation}

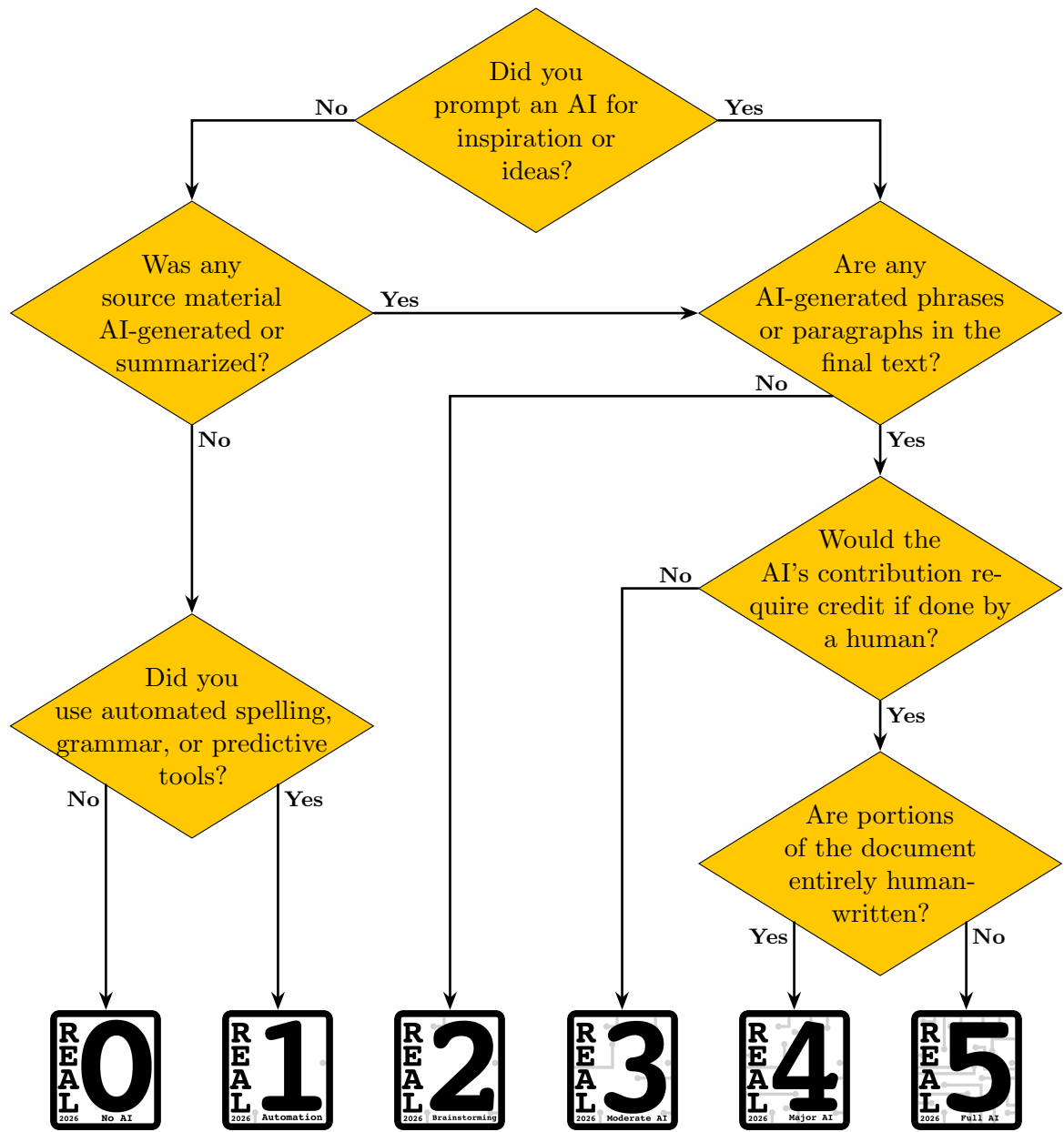
\begin{figure}[h!tbp]
  \centering
  \begin{tikzpicture}[x=5cm, y=4cm]
    \node[decision] (A) at (0, -0.3) {Did you \\ prompt an AI for inspiration or \\ ideas?};
    \node[decision] (B) at (-1,-1) {Was any \\ source material AI-generated or \\ summarized?};
    \node[decision, name path=C] (C) at (-1,-2.5) {Did you \\ use automated spelling, grammar, or predictive \\ tools?};
    \node[decision] (D) at (1,-1) {Are any \\ AI-generated phrases or paragraphs in the \\ final text?};
    \node[decision] (E) at (1,-2) {Would the \\ AI's contribution require credit if done by \\ a human?};
    \node[decision, name path=F] (F) at (1,-3) {Are portions \\ of the document entirely human-\\written?};
    
    \node[rating] (R0) at (-1.25,-3.75) {\ratingimg{images/0.png}};
    \node[rating] (R1) at (-0.75,-3.75) {\ratingimg{images/1.png}};
    \node[rating] (R2) at (-0.25,-3.75) {\ratingimg{images/2.png}};
    \node[rating] (R3) at (0.25,-3.75) {\ratingimg{images/3.png}};
    \node[rating] (R4) at (0.75,-3.75) {\ratingimg{images/4.png}};
    \node[rating] (R5) at (1.25,-3.75) {\ratingimg{images/5.png}};
    
    \draw[arrow] (A) -| node[label, pos=0, anchor=south east] {No} (B);
    \draw[arrow] (A) -| node[label, pos=0, anchor=south west] {Yes} (D);
    \draw[arrow] (B) -- node[label, pos=0, anchor=south west] {Yes} (D);
    \draw[arrow] (B) -- node[label, pos=0, anchor=north west] {No} (C);
    \path[name path=V0] (C -| R0) -- (R0);
    \draw[arrow, name intersections={of=C and V0}] (intersection-1) -| node[label, pos=0, anchor=north east] {No} (R0);
    \path[name path=V1] (C -| R1) -- (R1);
    \draw[arrow, name intersections={of=C and V1}] (intersection-1) -| node[label, pos=0, anchor=north west] {Yes} (R1);
    \draw[arrow] (D.240) -| node[label, pos=0.05, anchor=south east] {No} (R2);
    \draw[arrow] (D) -- node[label, pos=0, anchor=north west] {Yes} (E);
    \draw[arrow] (E) -- node[label, pos=0, anchor=north west] {Yes} (F);
    \draw[arrow] (E) -| node[label, pos=0, anchor=south east] {No} (R3);
    \path[name path=V4] (F -| R4) -- (R4);
    \draw[arrow, name intersections={of=F and V4}] (intersection-1) -| node[label, pos=0, anchor=north east] {Yes} (R4);
    \path[name path=V5] (F -| R5) -- (R5);
    \draw[arrow, name intersections={of=F and V5}] (intersection-1) -| node[label, pos=0, anchor=north west] {No} (R5);
  \end{tikzpicture}
  \caption{Text evaluation flowchart}\label{fig:text}
\end{figure}

Text evaluation addresses content copied directly from language models, significant co-authorship modifications, structural organization, and automated proofreading. This modality is comprehensive, covering written prose, software codebases, and the foundational scripts, research, or lyrics utilized in audio and video production (\autoref{fig:text}).

\section{Case Studies in Academic Research}

In academic contexts, the REAL Rating provides a framework for transparency in data analysis and manuscript preparation. While traditional authorship guidelines require significant human contribution, the inclusion of AI for literature summarization (Level 2) or code generation in research software (Level 3/4) is increasingly common. Applying the REAL Rating to research papers and datasets allows peer reviewers and the public to distinguish between human-derived conclusions and those influenced by algorithmic processing.

To illustrate how scholarly workflows align with the REAL framework, we consider several hypothetical scenarios that can serve as a baseline for consistent self-assessment.

\begin{description}

\item[Level 0 (No AI):] The investigators handle every stage manually, from data collection and analysis to manuscript preparation. Literature reviews rely solely on manual searching and reading, and the final paper is typeset in \LaTeX\ without any automated assistants or generative tools.

\item[Level 1 (Automation):] The researchers use standard word processors with basic spelling and grammar checks enabled. An automated citation manager handles the bibliography formatting, but the intellectual narrative and core content remain entirely human-driven.

\item[Level 2 (Brainstorming):] During the exploratory phase, the team uses an LLM to summarize research gaps within a specific set of literature. These summaries help narrow the thesis topic, but all subsequent experiments, data interpretation, and writing are done independently by the humans.

\item[Level 3 (Moderate AI):] While developing software to visualize experimental data, the researchers use an AI assistant to generate boilerplate code for UI components. The core algorithms and analytical logic are human-written, but the final tool includes AI-generated utility code. Alternatively, some portions of writing in the final manuscript are edited by an LLM, for example with the prompt ``lightly edit this paragraph for spelling, grammar, clarity and conciseness.'' Importantly, these edits must not significantly change the meaning or content of the writing.

\item[Level 4 (Major AI):] AI models act as active collaborators, generating significant portions of the data processing scripts and drafting initial results based on raw inputs. The AI's contribution mirrors that of a human co-author in scale, requiring intensive validation and editing from the human team.

\item[Level 5 (Full AI):] An automated agent produces a review paper by performing the literature search, data extraction, and meta-analysis before writing the entire manuscript. Human involvement is limited to setting the initial parameters and checking the final output for submission formatting.

\end{description}

Standardizing these disclosures gives journals a clear framework for transparency. More importantly, it simplifies the reviewer's job: once the level of AI involvement is out in the open, they can focus on the research itself instead of trying to reverse-engineer the production process.

\section{Conclusion}

The REAL Rating framework provides a standardized method for creators and organizations to report the extent of AI involvement in their media. By replacing binary labels with a six-tier criteria-based evaluation system, it allows for the precise disclosure of human-AI collaboration. This structured approach to provenance reporting offers audiences and institutions the specific context necessary to evaluate the origin and production methodology of digital content.

However, the REAL Rating framework possesses inherent limitations. Most notably, it operates as a self-reported system, relying on honest and accurate self-assessment of the creator. It does not replace cryptographic provenance standards like C2PA or automated detection tools, but serves as a complementary, user-facing semantic layer.

Furthermore, as an initiative developed by Real Good AI, an independent non-profit organization, the framework is distinctly positioned within the broader landscape of disclosure tools. Unlike proprietary detection tools seeking commercial software integration, or platform-specific labels designed primarily to mitigate corporate liability, the REAL Rating prioritizes transparent audience communication and public interest. As generative AI continues to integrate into professional and creative workflows, nuanced, independent disclosure mechanisms will be essential for maintaining digital trust.

\section*{Generative AI Disclosure}

\begin{wrapfigure}{R}{0.1\textwidth}
\raggedleft
\vspace{-1.5em}
\includegraphics[width=0.85\linewidth]{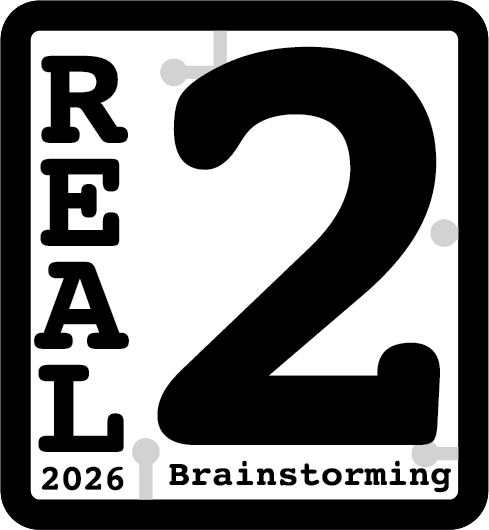} 
\end{wrapfigure}

Following the REAL framework described herein, this paper qualifies as \textbf{Level~2: Brainstorming}. AI was used as a collaborative tool for brainstorming and summarizing research.
\par

\bibliographystyle{plain}
\bibliography{real-rating}

\end{document}